\documentclass[runningheads]{llncs}
\usepackage[T1]{fontenc}
\usepackage{graphicx}
\usepackage{booktabs}
\usepackage{multirow}
\usepackage{amsmath}
\usepackage{listings}
\usepackage{amssymb}
\usepackage{subcaption}

\usepackage{hyperref}
\usepackage{color}

\begin{document}
\title{Large Language Models as Attribution Regularizers for Efficient Model Training}
\titlerunning{Large Language Models as Attribution Regularizers}
%
\author{Davor Vukadin\orcidID{0000-0003-3309-6718} \and
Marin Šilić\orcidID{0000-0002-4896-7689} \and
Goran Delač\orcidID{0000-0001-5315-8387}}
\authorrunning{D. Vukadin et al.}
%
\institute{University of Zagreb Faculty of Electrical Engineering and Computing, 3 Unska street, Zagreb, Croatia\\
\email{\{davor.vukadin, marin.silic, goran.delac\}@fer.hr}}
\maketitle              
\begin{abstract}
Large Language Models (LLMs) have demonstrated remarkable performance across diverse domains. However, effectively leveraging their vast knowledge for training smaller downstream models remains an open challenge, especially in domains like tabular data learning, where simpler models are often preferred due to interpretability and efficiency.

In this paper, we introduce a novel yet straightforward method for incorporating LLM-generated global task feature attributions into the training process of smaller networks. Specifically, we propose an attribution-matching regularization term that aligns the training dynamics of the smaller model with the insights provided by the LLM. By doing so, our approach yields superior performance in few-shot learning scenarios. Notably, our method requires only black-box API access to the LLM, making it easy to integrate into existing training pipelines with minimal computational overhead.

Furthermore, we demonstrate how this method can be used to address common issues in real-world datasets, such as skewness and bias. By integrating high-level knowledge from LLMs, our approach improves generalization, even when training data is limited or imbalanced. We validate its effectiveness through extensive experiments across multiple tasks, demonstrating improved learning efficiency and model robustness.

\end{abstract}
\section{Introduction}
The recent expansion in model parameters and training data for large language models (LLMs) has driven a significant breakthrough in natural language processing (NLP) \cite{DBLP:conf/nips/BrownMRSKDNSSAA20,DBLP:conf/naacl/DevlinCLT19,DBLP:journals/corr/abs-2203-15556,DBLP:conf/nips/VaswaniSPUJGKP17}. These models exhibit remarkable performance across various evaluation paradigms, such as zero-shot \cite{DBLP:conf/nips/KojimaGRMI22} and few-shot inference, leveraging in-context learning \cite{DBLP:conf/emnlp/MinLHALHZ22,DBLP:conf/nips/Wei0SBIXCLZ22}. This capability stems from the extensive text corpora used for training, which embed rich prior knowledge into LLMs, allowing them to approximate expert knowledge across diverse domains. Although their strong performance and generalization capabilities have been successfully extended to other modalities, such as images \cite{DBLP:conf/nips/LiuLWL23a} and speech \cite{DBLP:journals/corr/abs-2412-09596}, their application in tabular learning settings remains limited.

Several challenges hinder the adoption of LLMs for tabular data tasks. Firstly, their large parameter counts demand substantial computational resources, typically reliant on GPUs, which significantly increases operational costs. Secondly, tabular learning is often employed in domains where transparency and interpretability are critical, such as healthcare and finance. In these fields, simpler and more interpretable models, such as logistic regression or decision trees, are often preferred. Although techniques exist to enhance LLM interpretability \cite{DBLP:conf/icml/AchtibatHDJWLS24,leemann2025attention}, they currently fall short compared to the inherent explainability of simpler models.

In this paper, we investigate the potential of utilizing LLMs as training regularizers to enhance few-shot learning performance and improve generalization, especially in scenarios with skewed or biased training data. Specifically, we propose a method called \textbf{L}arge Language Model \textbf{A}ttribution \textbf{A}ligned \textbf{T}raining (\textbf{LAAT}), which introduces an attribution-matching regularization term that aligns the local, feature-wise explanations of smaller models with the global, task-specific explanations generated by LLMs. This approach harnesses the strong generalization capabilities of LLMs while preserving the efficiency and transparency of smaller models. We share our code at: \url{https://github.com/davor10105/laat}

\section{Related Work}
\subsection{Standard Machine Learning Approaches}
Inspired by the success of deep learning in other domains, numerous efforts have sought to apply self-supervised learning to tabular data to develop transfer learning-ready models. These approaches include masked feature prediction \cite{DBLP:conf/aaai/ArikP21,DBLP:journals/corr/abs-2206-08564}, feature corruption correction \cite{DBLP:conf/iclr/BahriJTM22,DBLP:conf/nips/YoonZJS20}, and contrastive pre-training \cite{DBLP:journals/corr/abs-2106-01342}. However, comparative studies indicate that gradient-boosted tree ensembles still outperform these methods \cite{DBLP:conf/nips/GrinsztajnOV22,DBLP:journals/inffus/Shwartz-ZivA22}. More recently, Nam et al. \cite{DBLP:conf/iclr/NamTLLS23} introduced Self-generated Tasks from UNlabeled Tables (STUNT), leveraging self-generated few-shot tasks for tabular learning, though its reliance on large unlabeled datasets may limit practical applicability. Additionally, Hollmann et al. \cite{DBLP:journals/nature/HollmannMPKKHSH25} proposed the Tabular Prior-data Fitted Network (TabPFN), a tabular foundation model pre-trained on millions of synthetic datasets.

\subsection{Large Language Models in Tabular Learning}
Most approaches integrating large language models (LLMs) into tabular learning rely on encoding task and feature descriptions in natural language, serializing the data, and leveraging LLMs for inference—either through in-context learning \cite{DBLP:journals/corr/abs-2304-13188} or additional fine-tuning \cite{DBLP:conf/nips/DinhZZLGRSP022,DBLP:conf/aistats/HegselmannBLA0S23,DBLP:conf/ijcai/0010GX024}. However, these methods face significant drawbacks, including the high cost of LLM inference for individual samples and the computational demands of fine-tuning.

In sensitive domains such as medicine or finance \cite{DBLP:journals/npjdm/ShickWKWDPSD24}, where transparency is critical, the opaque decision-making of LLMs is less desirable than traditional, smaller models. Alternative approaches involve using LLMs to generate synthetic examples to augment existing datasets, employing both in-context learning and fine-tuning \cite{DBLP:conf/icml/SeedatHBS24,DBLP:journals/corr/abs-2302-02041,DBLP:conf/emnlp/ZhangWYJL23}. However, these methods inherit the same scalability issues, particularly when dealing with high-dimensional datasets, where generating sufficiently large datasets becomes computationally expensive.

Recently, Han et al. \cite{DBLP:conf/icml/0001YAP24} introduced FeatLLM, a novel approach that utilizes LLMs as feature engineers. Instead of directly performing inference, FeatLLM employs code-generating LLMs to create preprocessing functions that transform the original dataset into a more suitable representation for few-shot classification. This method implements an ensemble classifier to combine insights from multiple feature transformations, improving robustness and classification accuracy. FeatLLM significantly reduces resource requirements by relying solely on pre-trained LLMs with API-level access. Moreover, FeatLLM outperforms existing fine-tuned and in-context learning approaches while maintaining lower computational costs. However, even though FeatLLM achieves state-of-the-art performance on few-shot tabular classification problems, its many iterations of rule and preprocessing function generation incur significant costs. Furthermore, during preprocessing, FeatLLM produces only binary features, which may limit expressiveness compared to the original data.

\subsection{Explanation Guided Learning}
A growing line of research explores enhancing model behavior through additional supervision derived from explainable artificial intelligence (XAI) techniques. This field can be broadly categorized into local explanation-guided learning and global explanation-guided learning \cite{DBLP:journals/csur/GaoGJHYZ24}.

Local explanation guidance applies supervision signals or regularization terms to individual model explanations, steering learning at the sample level. This approach is more prevalent due to the extensive development of local explanation techniques, particularly in the image domain, such as Grad-CAM \cite{DBLP:journals/ijcv/SelvarajuCDVPB20}, Layer-wise Relevance Propagation (LRP) \cite{Bach2015}, and attention-based attributions \cite{DBLP:conf/acl/AbnarZ20}. Ross et al. \cite{DBLP:conf/ijcai/RossHD17} propose regularizing differentiable models by penalizing input gradients, aligning them with expert-defined attribution maps. Dharma et al. \cite{DBLP:journals/corr/abs-2108-10131} use object bounding boxes as explanation supervision signals. In text classification, several studies leverage per-sample human-annotated rationales \cite{DBLP:conf/emnlp/ChoiPYH20,DBLP:conf/kdd/KanchinadamWYF20,DBLP:journals/corr/abs-1908-06870}. Gao et al. \cite{DBLP:conf/kdd/GaoSBGH022} demonstrate the effectiveness of local explanation supervision under limited training data. However, a key limitation of this approach is its reliance on per-sample attribution annotations, which are often difficult and costly to obtain, particularly in expert-driven fields like medicine.

Global explanation guidance, in contrast, does not require instance-level attributions, instead offering a broader, more scalable approach to shaping model behavior. Liu et al. \cite{DBLP:conf/acl/LiuA19} reduce undesired biases by penalizing nonzero attributions on sensitive tokens. Erion et al. \cite{DBLP:journals/natmi/ErionJSLL21} aggregate local feature attributions via expected gradients to improve interpretability. Weinberger et al. \cite{DBLP:conf/nips/WeinbergerJL20} extract prior knowledge from multiple gene expression datasets to construct meta-features, training a deep global attribution model alongside a predictive model with a regularization loss. However, this method assumes the availability of additional datasets related to the problem, which may not always be feasible.

\section{Method}
 In contrast to other methods that utilize LLMs for tabular data prediction, we seek to minimize both the computational and price overhead of their use, while simultaneously still effectively using their generalization abilities and providing small, interpretable models that can be readily used in existing pipelines.

 \subsection{Formulation}

Given a trained binary classification model $ m_{\theta}: \mathbb{R}^{N} \rightarrow [0, 1]$ parametrized by $\theta$, an attribution produced by an attribution method $a$ for an input $\mathbf{x}$ is a vector $a(x) = (s_1, ..., s_n)$, where $s_i$ is the attribution score of the input feature $x_i$. We are interested in the expected value of the attribution scores over the entire dataset, given by $ \mathbf{s}_{\mathbb{E}} = \mathbb{E}_{\mathbf{x} \sim \mathcal{D}}[a(\mathbf{x})]$, where $\mathcal{D}$ represents the data distribution.

For certain datasets where the expected attribution follows an intuitive pattern that humans can interpret, we hypothesize that this expected value can be approximated using a large language model and thus serve as a valuable local attribution guide during training.

Taking a step back, given an untrained model along with a task description and feature descriptions, we query an LLM to generate importance scores for each feature, producing a vector $\mathbf{s}_{\text{LLM}}$. During model training, we then regularize the local attribution scores of the model to align with these LLM-derived scores. This regularization acts as a guiding signal, helping the model maintain behavior that aligns with intuitive, human-understandable reasoning.

The final model's loss function consists of two components: the standard binary cross-entropy loss and an attribution regularization term. The regularization term is the mean squared error between the normalized attribution scores and the normalized LLM-derived scores, weighted by $\gamma$. The overall loss is given as a weighted sum of these terms:

\begin{equation}
\mathcal{L}(\theta) = \frac{1}{n}\sum^n_{i=1}(\ell_{\text{BCE}}(m_{\mathbf{\theta}}(\mathbf{x}_i),\mathbf{y}_i) + \gamma \ell_{\text{MSE}}(\frac{a(\mathbf{x}_i)}{\|a(\mathbf{x}_i)\|}, \frac{\mathbf{s}_{\text{LLM}}}{\|\mathbf{s}_{\text{LLM}}\|}))
\end{equation}

Following Ross et al. \cite{DBLP:conf/ijcai/RossHD17}, we employ the input gradient as our chosen attribution method.

\subsection{LLM Prompting and Score Parsing}

To enable Large Language Models (LLMs) to generate meaningful feature attribution scores for guiding downstream models, we developed a structured prompting methodology. This approach ensures score accuracy and relevance through three key components.

\subsubsection{Task and Dataset Contextualization}

We embed task and dataset details within the prompt. Task descriptions succinctly define classification objectives and outcomes, e.g., \textit{"Predict whether this patient's breast cancer will reoccur. Yes or no?"}, following established methodologies \cite{DBLP:conf/icml/0001YAP24,DBLP:conf/aistats/HegselmannBLA0S23}. Feature descriptions clarify dataset attributes, e.g., \textit{"Age: The age of the patient at diagnosis."}. Categorical features are one-hot encoded with explicit descriptions for each category, enabling the LLM to assign distinct attribution scores per category rather than per feature. Unlike prior approaches that use LLMs for tabular data classification, our method avoids serializing dataset examples into the prompt, thereby reducing prompt length and computational cost.

\subsubsection{Score Generation Protocol}

We instruct the LLM to assign integer scores between $-10$ and $10$, establishing a standardized feature importance scale. Using chain-of-thought prompting \cite{DBLP:conf/nips/Wei0SBIXCLZ22}, we ensure explicit reasoning before score assignment, enhancing interpretability. The full prompt template is provided in Prompt \ref{prompt:score_generation} in the Appendix.

\subsubsection{Score Extraction and Aggregation}

A secondary LLM instance extracts numerical scores from the primary LLM’s textual output via function calling, converting semi-structured responses into a standardized list format. To enhance stability, we generate scores multiple times ($\text{N}_{\text{estimates}}$) and compute their mean, yielding the final LLM-based feature attribution vector ($\mathbf{s}_{\text{LLM}}$).

This methodology provides a robust framework for leveraging LLM capabilities to generate reliable feature attribution scores that effectively inform downstream predictive models.

\section{Experiments}
We conducted a comprehensive evaluation of LAAT across diverse tabular datasets, examining its performance in few-shot learning contexts and scenarios where significant bias was present in the training data. Furthermore, we performed supplementary experiments to investigate the impact of various hyperparameter configurations on the proposed methodology. 

\subsection{Few-shot Learning}

While large volumes of data are readily available in many domains, expert-labeled data remains scarce in fields requiring specialized knowledge, such as medicine. To address this challenge, recent research has focused on enhancing the generalization capabilities of tabular classification models under minimal labeled data constraints \cite{DBLP:conf/icml/0001YAP24,DBLP:conf/aistats/HegselmannBLA0S23,DBLP:conf/iclr/NamTLLS23}. In this experiment, we assess the effectiveness of our proposed approach in leveraging LLM-derived knowledge as a guiding signal during extremely low-shot training scenarios.

We evaluate our approach on ten publicly available binary classification datasets:
\begin{itemize}
\item \textbf{adult} \cite{adult_2} - predicting whether an individual earns over \$50,000 annually
\item \textbf{bank} \cite{bank_marketing_222} - predicting whether a client will subscribe to a term deposit
\item \textbf{bodyfat} \cite{Penrose1985} - predicting whether an individual's body fat percentage exceeds the mean
\item \textbf{breast-ljub} \cite{breast_cancer_14} - predicting whether a patient's breast cancer will reoccur
\item \textbf{cdc-diabetes} \cite{cdc_diabetes} - predicting whether an individual has diabetes
\item \textbf{contraceptive} \cite{contraceptive} - predicting whether an individual uses contraception
\item \textbf{diabetes} \cite{diabetes_34} - predicting whether an individual has diabetes
\item \textbf{electricity} \cite{harries1999splice} - predicting the price change of electricity (up or down) in New South Wales
\item \textbf{indian-liver} \cite{indian_liver} - predict whether a patient has liver disease
\item \textbf{myocardial} \cite{myocardial_infarction_complications_579} - predicting whether the myocardial infarction complications data for an individual shows chronic heart failure
\end{itemize}
The datasets vary in size and complexity, and their additional basic information is outlined in Table \ref{tab:basic_dataset} in the Appendix.

\subsubsection{Baselines} We compare our proposed method with several baselines: logistic regression (LR), 2-layer MLP with ReLU activation and 100 hidden units (MLP), random forest (RF), XGBoost (XGB) \cite{DBLP:conf/kdd/ChenG16}, CatBoost \cite{DBLP:journals/corr/abs-1810-11363}, TabPFN \cite{DBLP:journals/nature/HollmannMPKKHSH25} and FeatLLM \cite{DBLP:conf/icml/0001YAP24}.

\subsubsection{Implementation Details} We employ three distinct foundation models as our importance score estimators: Llama 3.3 70B ($\text{LLa}_{\text{3.3}}$) \cite{DBLP:journals/corr/abs-2407-21783}, Gemini 2.0 Flash ($\text{Gem}_{\text{2.0}}$) \cite{DBLP:journals/corr/abs-2403-05530} and GPT-4o-mini ($\text{G}_{\text{4om}}$) \cite{DBLP:journals/corr/abs-2410-21276}. As simple downstream models, we utilized logistic regression and a two-layer multilayer perceptron (MLP) with ReLU activation and 100 hidden units. Optimization was performed using the Adam optimizer \cite{DBLP:journals/corr/KingmaB14} with a learning rate of $1 \times 10^{-2}$ and no weight decay, leveraging LAAT’s internal regularization via the $\gamma$ factor, which was set to 100. Additionally, the number of importance score estimations for LAAT was set to 5. LAAT was executed without early stopping. Data preprocessing involved one-hot encoding categorical variables and standardizing numerical features using z-score normalization. Feature descriptions were derived from dataset repository metadata or original publications, whereas task descriptions were formulated based on prior research on tabular data classification using LLMs \cite{DBLP:conf/icml/0001YAP24,DBLP:conf/aistats/HegselmannBLA0S23}. As prompt engineering is not the focus of our work, we leave the exploration of alternative prompting strategies to future research. Exact task descriptions used can be found in Table \ref{tab:task_descriptions} in the Appendix.

For traditional machine learning models (logistic regression, MLP, random forest, XGBoost, and CatBoost), we performed hyperparameter optimization through grid search with 5-fold cross-validation, with the exception of 1-shot learning scenarios where default scikit-learn \cite{scikit-learn} hyperparameters were applied. The complete baseline hyperparameter search spaces are detailed in Table \ref{tab:hyperparameters} in the Appendix. TabPFN and FeatLLM were implemented with their respective default parameter configurations as described in the original publications. Additionally, FeatLLM incorporated early stopping mechanisms in the 5-shot and 10-shot experimental settings. Due to the inherent complexity of generating multiple conditions and preprocessing functions in the FeatLLM method, Llama 3.3 exhibited inconsistencies in producing valid outputs. As a result, we excluded it from the FeatLLM experiments.

All models were trained using k-shot examples as the training set, while the remaining data served as the test set. To ensure statistical robustness, we conducted twenty independent experimental runs for each model. The significance of the difference in the mean values of the total scores was verified using the Wilcoxon signed-rank test with $p=0.05$.

In the following sections, we present our experimental results and provide additional analyses on LAAT’s robustness to noise, as well as an examination of how LAAT influences the training loss landscape through the lens of the bias-variance tradeoff.

\subsubsection{Results}

The results of our experiments, averaged over twenty repetitions, are presented in Tables \ref{tab:fewshot_baselines} (for non-LLM methods) and \ref{tab:fewshot_llm} (for LLM-based methods). As demonstrated, LAAT models consistently rank among the top-performing methods, securing the highest or shared highest scores in 28 out of 30 experiments. Notably, the highest-performing variants of LAAT were $\text{LAAT}^{MLP}_{G4om}$ and $\text{LAAT}^{LR}_{Gem2.0}$, securing 12 and 21 top or shared top scores respectively. Furthermore, the LAAT-trained models statistically significantly outperformed their non-LAAT counterparts in 24, 28, and 27 out of 30 experiments using logistic regression with Llama 3.3 70B, Gemini 2.0 Flash, and GPT-4o-mini as scoring models, respectively. For MLP, LAAT models significantly exceeded the performance of vanilla MLP in 23, 28, and 29 experiments, respectively. These results underscore the positive impact of LLM attribution alignment on the generalization capabilities of even simple models, achieving significant improvements without the need for extensive hyperparameter tuning, which is often required for other approaches.

\begin{table}
\centering
\caption{ROC AUC scores of baseline models on the few-shot experiments. Best scores for each dataset, across both LLM and non-LLM approaches, are emphasized in \textbf{bold}. Multiple bolded values indicate that their differences were not statistically significant according to the Wilcoxon signed-rank test at $p=0.05$.}\label{tab:fewshot_baselines}
\begin{tabular}{l|l||l|l|l|l|l|l}
\hline
\textbf{Dataset} & \textbf{Shot} &  LR & MLP & RF & XGB & CatBoost & TabPFN  \\
\hline
\hline
\multirow[t]{3}{*}{adult} & 1 & $58.9_{19.5}$ & $57.9_{14.1}$ & $67.9_{11.8}$ & $50.0_{0.0}$ & $64.7_{14.5}$ & $64.9_{16.3}$ \\
 & 5 & $76.7_{6.6}$ & $72.8_{7.0}$ & $73.8_{9.2}$ & $63.9_{11.9}$ & $78.1_{4.8}$ & $78.1_{4.7}$ \\
 & 10 & $81.0_{3.4}$ & $78.4_{4.8}$ & $76.4_{8.1}$ & $77.4_{3.3}$ & $82.3_{3.5}$ & $82.2_{3.5}$ \\
\cline{1-8}
\multirow[t]{3}{*}{bank} & 1 & $51.2_{12.5}$ & $55.5_{10.7}$ & $56.0_{9.1}$ & $50.0_{0.0}$ & $56.8_{9.2}$ & $56.2_{8.7}$ \\
 & 5 & $63.9_{8.7}$ & $64.6_{7.8}$ & $65.7_{8.4}$ & $56.4_{9.3}$ & $68.2_{8.1}$ & $67.5_{9.1}$ \\
 & 10 & $66.7_{3.9}$ & $67.0_{6.3}$ & $69.1_{6.8}$ & $71.4_{6.9}$ & $74.7_{5.1}$ & $74.5_{7.6}$ \\
\cline{1-8}
\multirow[t]{3}{*}{bodyfat} & 1 & $62.7_{22.4}$ & $69.0_{14.2}$ & $64.4_{19.1}$ & $50.0_{0.0}$ & $67.1_{15.6}$ & $68.6_{13.7}$ \\
 & 5 & $78.3_{9.6}$ & $78.7_{10.1}$ & $74.8_{8.0}$ & $64.3_{14.1}$ & $75.2_{11.9}$ & $78.0_{10.4}$ \\
 & 10 & $84.4_{3.1}$ & $84.3_{3.7}$ & $79.4_{6.9}$ & $80.2_{5.0}$ & $83.6_{2.6}$ & $86.2_{3.9}$ \\
\cline{1-8}
\multirow[t]{3}{*}{breast-ljub} & 1 & $55.1_{9.7}$ & $56.7_{12.7}$ & $55.8_{7.2}$ & $50.0_{0.0}$ & $52.8_{9.1}$ & $53.1_{9.8}$ \\
 & 5 & $58.4_{10.6}$ & $57.9_{6.5}$ & $61.6_{8.3}$ & $55.0_{7.8}$ & $60.2_{10.0}$ & $59.9_{9.8}$ \\
 & 10 & $61.6_{5.9}$ & $61.2_{5.6}$ & $60.9_{8.2}$ & $61.6_{8.1}$ & $64.6_{6.5}$ & $66.4_{5.6}$ \\
\cline{1-8}
\multirow[t]{3}{*}{cdc-diabetes} & 1 & $58.0_{12.9}$ & $62.4_{11.9}$ & $59.1_{15.6}$ & $50.0_{0.0}$ & $59.7_{12.5}$ & $64.1_{11.4}$ \\
 & 5 & $69.3_{5.9}$ & $63.3_{11.5}$ & $66.0_{8.4}$ & $60.6_{5.7}$ & $73.0_{4.5}$ & $72.1_{4.0}$ \\
 & 10 & $71.8_{4.3}$ & $67.0_{12.1}$ & $66.7_{8.5}$ & $69.8_{5.6}$ & $75.5_{3.0}$ & $73.3_{4.1}$ \\
\cline{1-8}
\multirow[t]{3}{*}{contraceptive} & 1 & $50.9_{4.9}$ & $52.4_{5.0}$ & $54.4_{5.7}$ & $50.0_{0.0}$ & $51.9_{3.7}$ & $51.3_{5.6}$ \\
 & 5 & $54.2_{4.9}$ & $53.0_{5.2}$ & $54.5_{6.7}$ & $52.2_{4.9}$ & $56.0_{6.8}$ & $56.9_{5.9}$ \\
 & 10 & $56.5_{3.9}$ & $53.6_{6.1}$ & $57.3_{3.9}$ & $57.5_{4.9}$ & $59.2_{4.6}$ & $59.4_{3.8}$ \\
\cline{1-8}
\multirow[t]{3}{*}{diabetes} & 1 & $53.9_{13.6}$ & $58.3_{15.6}$ & $63.5_{7.9}$ & $50.0_{0.0}$ & $59.2_{8.3}$ & $58.5_{10.1}$ \\
 & 5 & $70.6_{6.4}$ & $64.2_{9.8}$ & $66.0_{7.4}$ & $61.6_{8.7}$ & $72.6_{5.1}$ & $72.9_{5.3}$ \\
 & 10 & $73.8_{7.9}$ & $72.7_{5.8}$ & $68.8_{7.2}$ & $70.9_{7.4}$ & $75.3_{5.1}$ & $75.8_{5.9}$ \\
\cline{1-8}
\multirow[t]{3}{*}{electricity} & 1 & $51.5_{12.3}$ & $54.1_{13.1}$ & $58.5_{9.5}$ & $50.0_{0.0}$ & $60.2_{11.8}$ & $58.9_{10.2}$ \\
 & 5 & $65.2_{7.9}$ & $69.0_{6.4}$ & $63.6_{8.3}$ & $58.3_{9.9}$ & $65.4_{8.1}$ & $66.8_{7.9}$ \\
 & 10 & $\mathbf{72.6_{4.5}}$ & $68.0_{8.8}$ & $70.7_{4.4}$ & $70.5_{7.2}$ & $\mathbf{74.5_{2.2}}$ & $\mathbf{73.2_{5.1}}$ \\
\cline{1-8}
\multirow[t]{3}{*}{indian-liver} & 1 & $54.3_{14.7}$ & $55.6_{14.8}$ & $60.0_{11.6}$ & $50.0_{0.0}$ & $58.5_{11.3}$ & $61.1_{11.8}$ \\
 & 5 & $61.3_{10.5}$ & $60.4_{12.3}$ & $60.3_{10.3}$ & $58.9_{7.1}$ & $67.7_{4.5}$ & $63.4_{6.3}$ \\
 & 10 & $67.4_{7.3}$ & $65.9_{4.7}$ & $62.1_{6.8}$ & $62.7_{7.0}$ & $70.6_{3.2}$ & $69.4_{4.2}$ \\
\cline{1-8}
\multirow[t]{3}{*}{myocardial} & 1 & $50.5_{5.7}$ & $52.1_{3.0}$ & $51.9_{5.8}$ & $50.0_{0.0}$ & $50.8_{4.0}$ & $50.8_{4.1}$ \\
 & 5 & $55.3_{5.2}$ & $53.3_{7.6}$ & $52.3_{6.8}$ & $50.9_{3.4}$ & $53.3_{6.0}$ & $54.6_{5.4}$ \\
 & 10 & $59.3_{5.4}$ & $58.5_{5.2}$ & $53.4_{5.5}$ & $52.3_{4.4}$ & $58.7_{5.0}$ & $61.0_{5.8}$ \\
\cline{1-8}
\hline
\end{tabular}
\end{table}

\begin{table}
\centering
\caption{ROC AUC scores of LLM-based models on the few-shot experiments. Best scores for each dataset, across both LLM and non-LLM approaches, are emphasized in \textbf{bold}. Multiple bolded values indicate that their differences were not statistically significant according to the Wilcoxon signed-rank test at $p=0.05$.}\label{tab:fewshot_llm}
\begin{tabular}{l|l||ll|lll|lll}
\hline
\multicolumn{2}{c||}{\textbf{Method}} & \multicolumn{2}{c|}{FeatLLM} & \multicolumn{6}{c}{LAAT} \\
\hline
\multicolumn{2}{c||}{\textbf{Model}} & \multicolumn{2}{c|}{Ensemble} & \multicolumn{3}{c|}{LR} & \multicolumn{3}{c}{MLP} \\
\hline
\textbf{Dataset} & \textbf{Shot} & $Gem_{2.0}$ & $G_{4om}$ & $LLa_{3.3}$ & $Gem_{2.0}$ & $G_{4om}$ & $LLa_{3.3}$ & $Gem_{2.0}$ & $G_{4om}$  \\
\hline
\hline
\multirow[t]{3}{*}{adult} & 1 & $\mathbf{81.2_{6.1}}$ & $\mathbf{80.4_{4.7}}$ & $68.9_{11.6}$ & $67.0_{13.1}$ & $70.5_{10.3}$ & $70.5_{8.6}$ & $71.4_{9.9}$ & $72.8_{7.6}$ \\
 & 5 & $\mathbf{85.3_{2.5}}$ & $81.1_{4.2}$ & $79.2_{4.5}$ & $81.5_{4.6}$ & $77.9_{4.2}$ & $76.3_{4.8}$ & $78.7_{5.2}$ & $77.5_{4.5}$ \\
 & 10 & $\mathbf{84.4_{3.0}}$ & $78.5_{5.1}$ & $82.4_{3.4}$ & $\mathbf{85.2_{3.4}}$ & $81.0_{3.2}$ & $79.1_{4.8}$ & $82.1_{4.4}$ & $78.6_{3.7}$ \\
\cline{1-10}
\multirow[t]{3}{*}{bank} & 1 & $71.4_{3.3}$ & $66.3_{3.0}$ & $61.3_{8.7}$ & $\mathbf{76.2_{10.3}}$ & $\mathbf{77.2_{7.5}}$ & $58.1_{9.4}$ & $72.5_{10.1}$ & $74.4_{7.8}$ \\
 & 5 & $71.9_{5.0}$ & $66.3_{4.3}$ & $65.6_{2.2}$ & $\mathbf{85.6_{2.0}}$ & $83.8_{0.8}$ & $64.4_{4.9}$ & $82.0_{5.0}$ & $83.4_{2.6}$ \\
 & 10 & $70.3_{4.0}$ & $64.2_{5.1}$ & $64.7_{1.4}$ & $\mathbf{86.0_{0.9}}$ & $84.7_{1.0}$ & $65.6_{4.5}$ & $81.9_{3.9}$ & $84.4_{2.1}$ \\
\cline{1-10}
\multirow[t]{3}{*}{bodyfat} & 1 & $67.6_{11.5}$ & $76.0_{7.4}$ & $88.2_{1.6}$ & $\mathbf{89.6_{0.8}}$ & $89.0_{1.2}$ & $84.9_{6.1}$ & $85.6_{6.6}$ & $85.5_{6.2}$ \\
 & 5 & $82.6_{2.4}$ & $81.6_{2.4}$ & $89.0_{0.5}$ & $\mathbf{90.2_{0.4}}$ & $89.7_{0.4}$ & $89.0_{0.5}$ & $\mathbf{90.2_{0.5}}$ & $89.7_{0.5}$ \\
 & 10 & $82.6_{2.4}$ & $78.8_{7.1}$ & $89.0_{0.6}$ & $\mathbf{90.2_{0.5}}$ & $89.7_{0.6}$ & $89.0_{0.7}$ & $\mathbf{90.2_{0.6}}$ & $89.8_{0.7}$ \\
\cline{1-10}
\multirow[t]{3}{*}{breast-ljub} & 1 & $60.9_{6.9}$ & $61.5_{7.9}$ & $74.0_{0.9}$ & $73.9_{1.1}$ & $\mathbf{74.4_{0.7}}$ & $72.7_{3.0}$ & $72.6_{2.7}$ & $72.6_{3.1}$ \\
 & 5 & $65.0_{7.1}$ & $61.4_{5.5}$ & $74.4_{0.8}$ & $\mathbf{74.7_{0.8}}$ & $74.4_{0.8}$ & $74.1_{1.0}$ & $\mathbf{74.4_{1.0}}$ & $74.0_{1.1}$ \\
 & 10 & $63.9_{6.8}$ & $61.0_{8.5}$ & $73.8_{1.1}$ & $\mathbf{74.2_{1.1}}$ & $73.7_{1.1}$ & $73.6_{1.3}$ & $73.8_{1.4}$ & $73.2_{1.4}$ \\
\cline{1-10}
\multirow[t]{3}{*}{cdc-diabetes} & 1 & $\mathbf{73.7_{2.8}}$ & $72.6_{2.9}$ & $72.5_{2.8}$ & $\mathbf{75.1_{2.7}}$ & $72.6_{2.1}$ & $70.8_{4.2}$ & $74.3_{2.9}$ & $72.1_{3.2}$ \\
 & 5 & $71.4_{11.1}$ & $75.3_{1.5}$ & $74.4_{1.9}$ & $\mathbf{78.1_{0.9}}$ & $75.1_{1.2}$ & $74.7_{2.2}$ & $\mathbf{78.2_{0.9}}$ & $75.6_{1.5}$ \\
 & 10 & $73.2_{3.3}$ & $73.4_{2.6}$ & $74.8_{1.3}$ & $\mathbf{78.5_{0.6}}$ & $75.7_{0.9}$ & $75.2_{1.4}$ & $\mathbf{78.6_{0.6}}$ & $76.3_{1.1}$ \\
\cline{1-10}
\multirow[t]{3}{*}{contraceptive} & 1 & $55.7_{4.5}$ & $55.6_{4.3}$ & $\mathbf{62.8_{0.8}}$ & $\mathbf{61.6_{2.4}}$ & $61.1_{1.3}$ & $\mathbf{63.1_{2.3}}$ & $61.2_{3.3}$ & $61.3_{2.2}$ \\
 & 5 & $53.4_{5.8}$ & $53.1_{5.5}$ & $62.7_{0.2}$ & $\mathbf{64.2_{0.7}}$ & $61.7_{0.1}$ & $62.9_{0.4}$ & $\mathbf{64.2_{1.4}}$ & $61.7_{0.3}$ \\
 & 10 & $54.7_{4.6}$ & $53.6_{4.0}$ & $62.6_{0.2}$ & $64.9_{0.4}$ & $61.6_{0.2}$ & $62.9_{0.4}$ & $\mathbf{65.2_{0.7}}$ & $61.7_{0.3}$ \\
\cline{1-10}
\multirow[t]{3}{*}{diabetes} & 1 & $75.6_{4.5}$ & $73.1_{3.7}$ & $72.8_{8.1}$ & $\mathbf{78.9_{2.9}}$ & $\mathbf{75.0_{8.0}}$ & $68.4_{9.8}$ & $69.8_{9.3}$ & $69.7_{9.4}$ \\
 & 5 & $73.1_{8.9}$ & $75.0_{2.0}$ & $78.7_{1.5}$ & $\mathbf{79.8_{0.4}}$ & $\mathbf{79.7_{0.6}}$ & $78.8_{1.5}$ & $\mathbf{79.8_{0.5}}$ & $\mathbf{79.7_{0.7}}$ \\
 & 10 & $70.9_{5.7}$ & $75.4_{2.3}$ & $79.1_{0.7}$ & $79.6_{0.3}$ & $79.7_{0.4}$ & $79.2_{0.7}$ & $\mathbf{79.7_{0.4}}$ & $\mathbf{79.8_{0.4}}$ \\
\cline{1-10}
\multirow[t]{3}{*}{electricity} & 1 & $63.9_{4.9}$ & $66.5_{5.7}$ & $65.5_{3.7}$ & $66.8_{3.4}$ & $\mathbf{73.3_{5.3}}$ & $61.4_{7.1}$ & $61.8_{8.2}$ & $65.4_{9.8}$ \\
 & 5 & $71.8_{4.1}$ & $67.6_{5.3}$ & $67.6_{0.6}$ & $68.3_{0.2}$ & $\mathbf{74.9_{1.5}}$ & $67.4_{0.7}$ & $68.2_{0.4}$ & $\mathbf{74.9_{1.6}}$ \\
 & 10 & $70.8_{3.8}$ & $65.5_{6.3}$ & $67.4_{0.4}$ & $68.4_{0.3}$ & $\mathbf{74.1_{2.0}}$ & $67.4_{0.5}$ & $68.3_{0.4}$ & $\mathbf{74.2_{2.0}}$ \\
\cline{1-10}
\multirow[t]{3}{*}{indian-liver} & 1 & $68.5_{4.1}$ & $67.6_{4.9}$ & $\mathbf{73.1_{1.6}}$ & $\mathbf{73.4_{1.1}}$ & $72.7_{1.7}$ & $72.0_{2.1}$ & $72.1_{1.7}$ & $71.4_{2.2}$ \\
 & 5 & $72.2_{1.7}$ & $72.7_{0.8}$ & $\mathbf{73.8_{1.0}}$ & $\mathbf{73.7_{0.9}}$ & $73.2_{1.0}$ & $\mathbf{73.7_{0.9}}$ & $73.5_{1.0}$ & $73.1_{0.9}$ \\
 & 10 & $70.3_{3.8}$ & $70.5_{2.5}$ & $\mathbf{74.0_{0.9}}$ & $\mathbf{73.9_{0.7}}$ & $73.3_{0.9}$ & $\mathbf{73.9_{0.9}}$ & $73.9_{1.0}$ & $73.4_{1.1}$ \\
\cline{1-10}
\multirow[t]{3}{*}{myocardial} & 1 & $59.0_{3.5}$ & $59.1_{4.4}$ & $54.8_{5.3}$ & $58.3_{6.3}$ & $57.9_{6.3}$ & $58.5_{2.9}$ & $\mathbf{63.1_{3.9}}$ & $61.7_{4.4}$ \\
 & 5 & $59.7_{5.0}$ & $60.6_{3.0}$ & $60.7_{3.7}$ & $\mathbf{65.9_{2.5}}$ & $64.9_{3.0}$ & $62.0_{1.8}$ & $\mathbf{66.1_{1.8}}$ & $64.4_{2.6}$ \\
 & 10 & $58.7_{2.8}$ & $60.7_{3.2}$ & $63.0_{2.7}$ & $\mathbf{67.2_{2.2}}$ & $66.5_{3.0}$ & $62.6_{2.4}$ & $\mathbf{66.7_{2.5}}$ & $65.9_{2.7}$ \\
\cline{1-10}
\hline
\end{tabular}
\end{table}

\begin{figure}
\centering
\includegraphics[width=\textwidth]{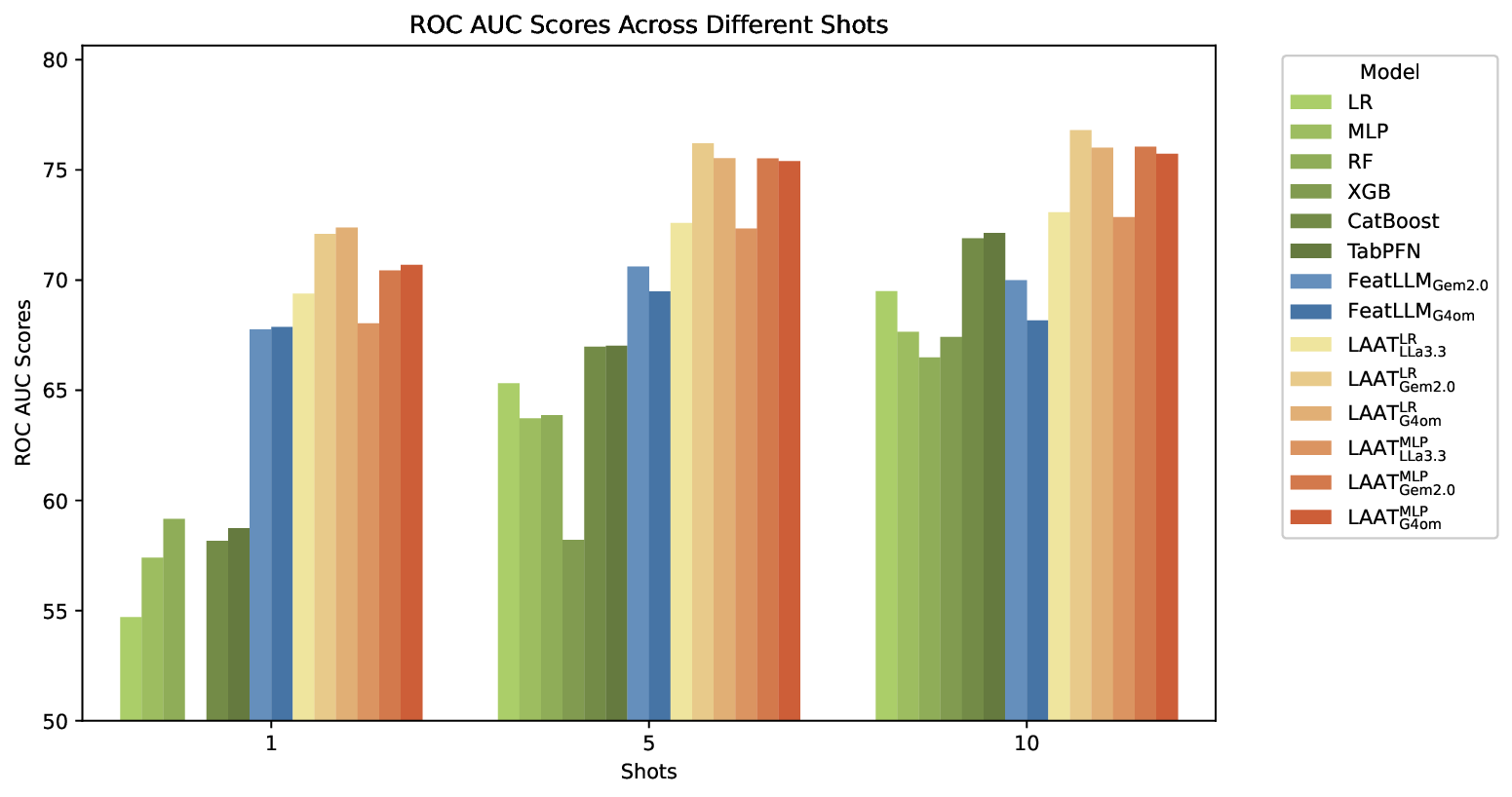}
\caption{ROC AUC scores for baseline models and LLM-based methods, averaged across all datasets. In the 1-shot setting, FeatLLM achieves the highest average performance. However, in subsequent shots, LAAT-based approaches, both Logistic Regression and MLP variants, outperform the baselines and FeatLLM.} \label{fig:fewshot}
\end{figure}

Figure \ref{fig:fewshot} illustrates the average ROC AUC performance across all datasets, highlighting the superior performance of LAAT model variants across all shot settings. While FeatLLM closely matches LAAT in the 1-shot scenario, its performance plateaus beyond this point. We hypothesize that this occurs for two reasons: first, FeatLLM generates binary features, which may lack the expressiveness of the continuous features present in the original dataset. Second, the serialized few-shot examples significantly increase the length of the input prompt, potentially reducing the effectiveness of the rule and preprocessing function generation procedure. This may occur because the expanded prompt causes the initial instruction and relevant data to become less prominent among the large number of tokens. In contrast, LAAT models demonstrate a substantial advantage in both the 5-shot and 10-shot settings, significantly outperforming all other methods. Among the scoring models evaluated, Gemini 2.0 Flash-Lite emerged as the top performer, with GPT-4o-mini following closely behind. Although Llama 3.3 70B lagged behind these two LLMs, it consistently matched or outperformed all baseline methods across all settings. In addition to outperforming FeatLLM, LAAT utilizes, on average, 79\% fewer input tokens and 60\% fewer output tokens, as shown in Table \ref{tab:token_comparison}, demonstrating significant conservation of computational resources.

\begin{table}
\centering
\caption{FeatLLM and LAAT average token count comparison. LAAT consumes significantly less input and output tokens, conserving computational resources.}\label{tab:token_comparison}
\begin{tabular}{c||cccccc|cc}
\hline
\textbf{Method} & \multicolumn{6}{c|}{FeatLLM} & \multicolumn{2}{c}{LAAT} \\
\hline
\textbf{k-shot} & \multicolumn{2}{c}{1} & \multicolumn{2}{c}{5} & \multicolumn{2}{c|}{10} & \multicolumn{2}{c}{N/A} \\
\hline
\textbf{Token Type} & Input & Output & Input & Output & Input & Output & Input & Output \\
\textbf{Count} & 37866 & 19730 & 51704 & 20067 & 62310 & 20282 & \textbf{10570} & \textbf{7990} \\
\hline
\end{tabular}
\end{table}

\subsubsection{Sensitivity to Importance Score Noise }

We assess the robustness of the LAAT method by introducing controlled noise into the importance scores provided by the LLM. Specifically, we define a noise ratio $\epsilon \in [0, 1]$ and compute the perturbed importance scores, $\mathbf{s}^{\text{noisy}}_{\text{LLM}}$, as a linear interpolation between the original scores and randomly generated scores, $\mathbf{s}_{\text{noise}}$, sampled from a uniform integer distribution in the range $[-10, 10]$:

\begin{equation}
\mathbf{s}^{\text{noisy}}_{\text{LLM}} = (1 - \epsilon) \mathbf{s}_{\text{LLM}} + \epsilon \mathbf{s}_{\text{noise}}   
\end{equation}

\begin{figure}
\centering
\includegraphics[width=0.55\textwidth]{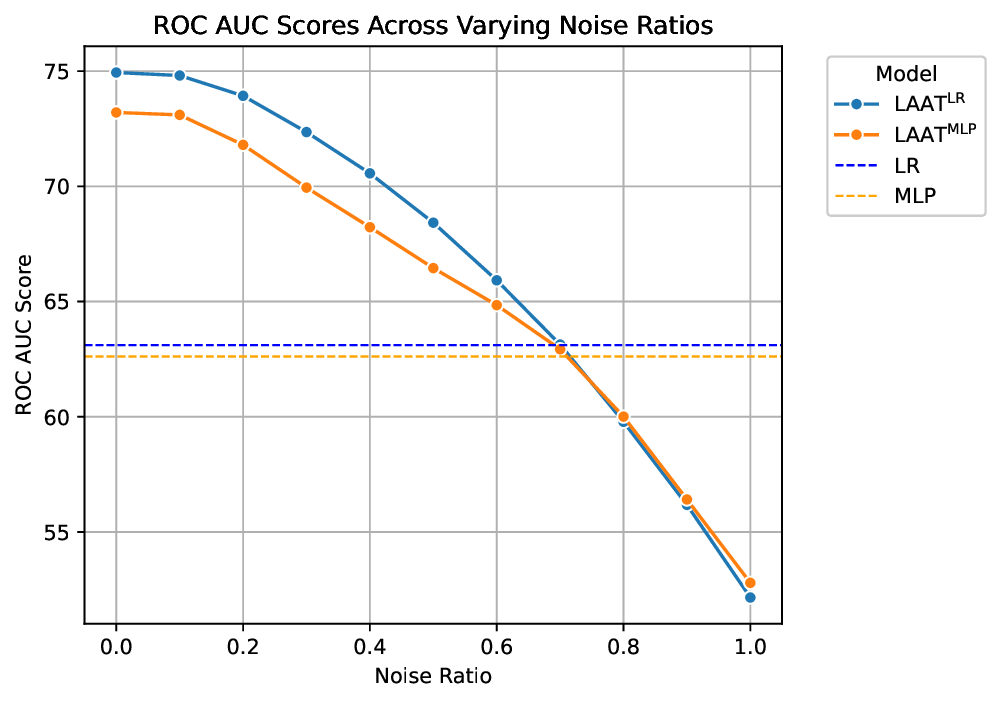}
\caption{LAAT robustness to noisy importance scores. Despite performance degradation with increasing noise $\epsilon$, LAAT variants outperform non-LAAT counterparts up to $\epsilon = 0.6$, demonstrating resilience to imperfect LLM scores.} \label{fig:noise_ratio}
\end{figure}

Figure \ref{fig:noise_ratio} presents the average performance across all shot scenarios for both LR and MLP LAAT variants using the GPT-4o mini model under varying noise conditions. As expected, the performance of LAAT declines as the noise ratio increases. However, even with noise ratios as high as 0.6, LAAT variants continue to outperform their non-LAAT counterparts. These results demonstrate LAAT's ability to enhance baseline model performance even when the LLM-derived importance scores are imperfect.

\subsubsection{Exploring the Loss Landscape}

To understand why LAAT consistently outperforms other methods, we analyze the training and test loss landscapes \cite{losslandscapes} of two logistic regression models: one trained with standard binary cross-entropy loss and another incorporating the attribution alignment loss with $\gamma = 100$. We also visualize their respective training trajectories. Each model was trained using Adam for 200 epochs without early stopping or weight decay, with only five training samples per class, while the remaining dataset was used as the test set.

\begin{figure}
    \centering
    \begin{subfigure}{0.8\textwidth}
        \centering
        \includegraphics[width=\linewidth]{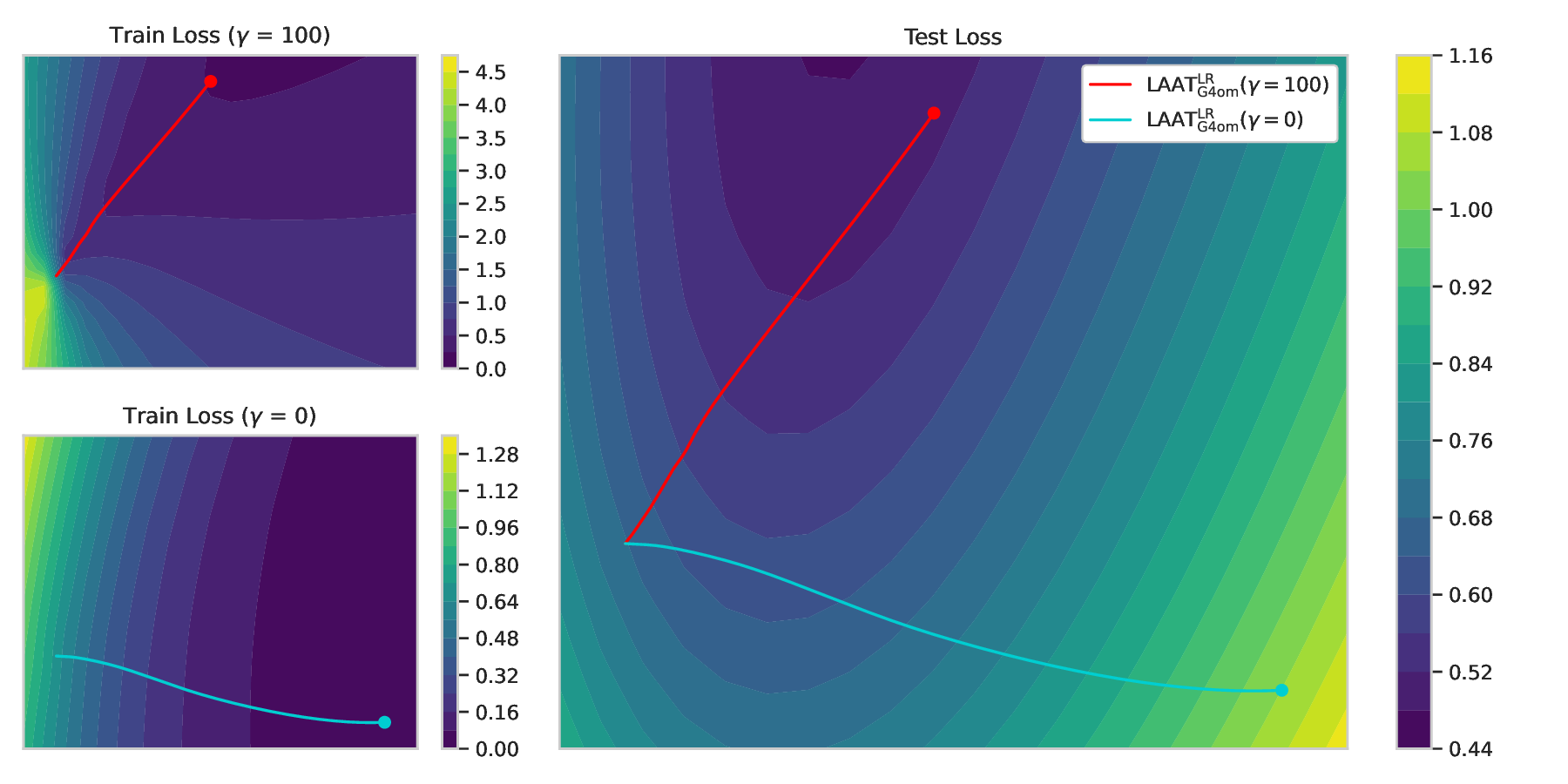}
        \caption{Loss landscapes - \textbf{adult} dataset.}
    \end{subfigure}
    \hfill
    \begin{subfigure}{0.8\textwidth}
        \centering
        \includegraphics[width=\linewidth]{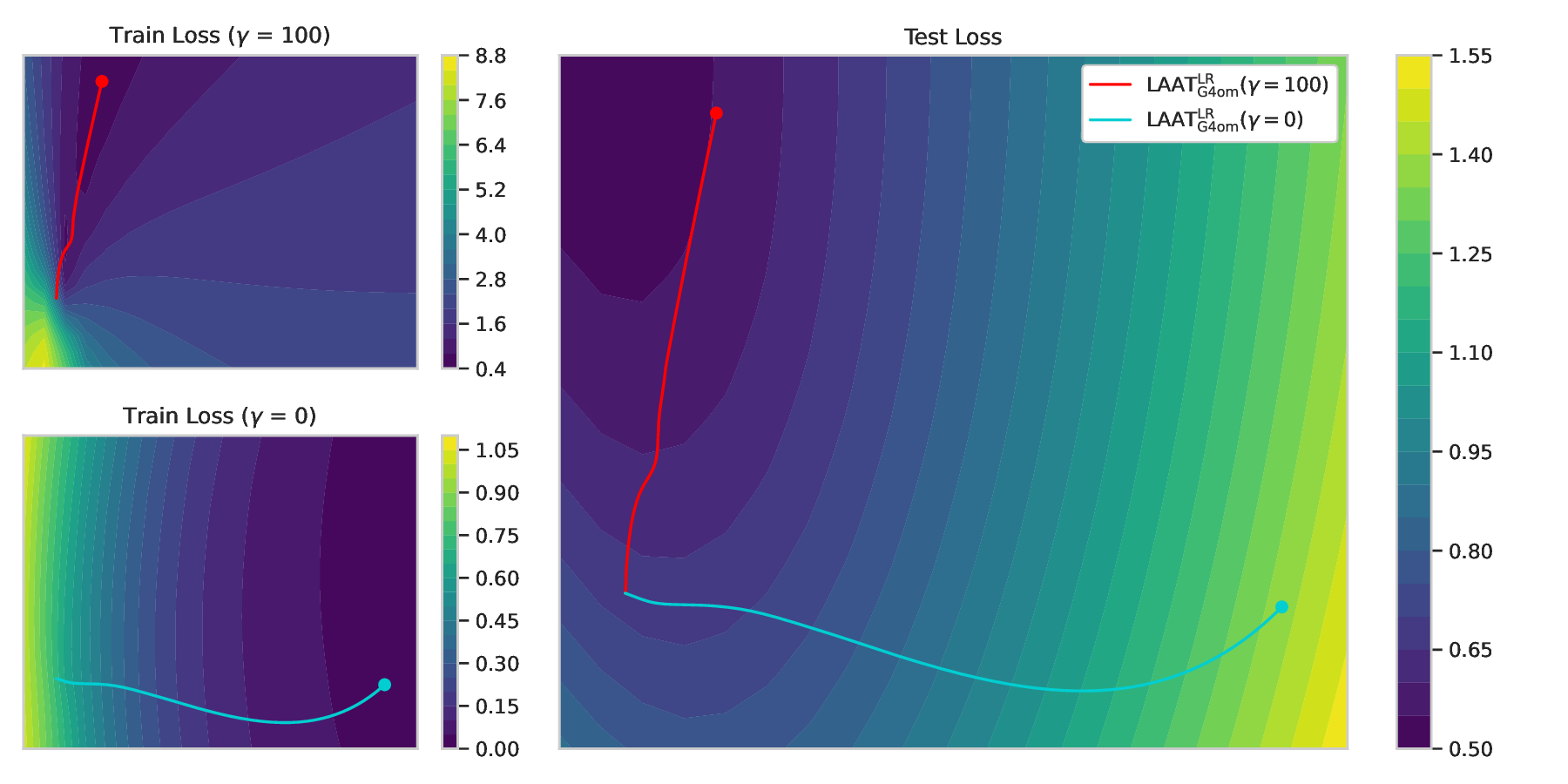}
        \caption{Loss landscapes - \textbf{bank} dataset.}
    \end{subfigure}
    \caption{Train and test loss landscapes for logistic regression on the \textbf{adult} and \textbf{bank} datasets, comparing models with attribution alignment loss ($\gamma = 100$, top left subfigure) and without it ($\gamma = 0$, bottom left subfigure). The addition of attribution alignment loss results in a closer match between training landscapes (left half) and the test loss landscape (right half), guiding the training process (colored lines ending with a point) towards minima that align well with testing minima, indicating improved generalization.}
    \label{fig:loss_landscapes_adult_bank}
\end{figure}

Figure \ref{fig:loss_landscapes_adult_bank} illustrates the results for the \textbf{adult} and \textbf{bank} datasets, while additional results for the \textbf{breast-ljub} and \textbf{myocardial} datasets are provided in the Appendix in Figure \ref{fig:loss_landscapes_breast_myocardial}.

A key observation from these visualizations is that the loss landscape of the standard model exhibits a substantial discrepancy between training and test loss surfaces. This suggests high variance, where the model overfits to the limited training data, learning feature importance patterns that do not generalize well. Conversely, introducing the attribution alignment loss alters the training loss landscape, shifting it closer to the test loss landscape. This shift can be interpreted as a form of inductive bias, where the model is nudged toward an attribution structure informed by the LLM’s prior knowledge.  

From a bias-variance perspective, LAAT's regularization constrains the model’s optimization path, discouraging reliance on spurious correlations and guiding it toward more stable, generalizable parameters. Secondly, the model becomes less sensitive to variations in small datasets, lowering the variance, which is preferable in low-data scenarios. This trade-off mitigates overfitting and enhances generalization, ultimately leading to improved predictive performance, as demonstrated in previous experiments.

\subsection{Learning on Biased Data}
Many real-world datasets exhibit varying degrees of bias, which can negatively impact machine learning models by introducing misleading correlations. For instance, a model trained on historical recruitment data may inherit biases related to gender and job roles \cite{Chen2023}. Similarly, systemic biases in criminal justice datasets often lead to strong correlations between demographic attributes and outcomes, potentially reflecting enforcement patterns rather than actual crime rates. In some cases, these confounding factors can be explicitly removed when they are demonstrably unrelated to the target classification. However, the challenge becomes more complex when the relationship is "soft"—i.e., a correlation exists, but its exact influence is difficult to quantify.

To evaluate the performance of baseline methods alongside our proposed LAAT method, we introduce artificial biases into the training portion of the medicine-related datasets used in the previous experiment while keeping the evaluation data distribution unchanged. This approach mimics real-world biases that may arise due to the way data is collected—for example, demographic skews in clinical trials or socioeconomic biases in electronic health records. By training models on biased distributions and assessing them on the original, unbiased test data, we simulate practical scenarios where spurious correlations may mislead models, allowing us to analyze their robustness to such biases. Specifically, we apply the following modifications:
\begin{itemize}
\item \textbf{bodyfat} - Exclude all individuals under the age of 50 with above-average body fat percentage, introducing a bias that overestimates body fat in elderly individuals.
\item \textbf{breast-ljub} - Exclude all patients under the age of 50 who experienced breast cancer recurrence, introducing a bias that overestimates recurrence in elderly patients.
\item \textbf{cdc-diabetes} - Exclude all women with diabetes and all men without diabetes, artificially inflating the apparent prevalence of diabetes among men.
\item \textbf{contraceptive} - Exclude all working women not using contraceptives and all non-working women using contraceptives, creating a spurious relationship between employment and contraceptive use.
\item \textbf{diabetes} - Remove all individuals under 50 years old who have diabetes, encouraging models to over-rely on age for diabetes prediction.
\item \textbf{indian-liver} - Remove all male patients with diagnosed liver disease, biasing models toward associating liver disease more strongly with female patients.
\item \textbf{myocardial} - Remove all female patients with a history of myocardial infarction, biasing models toward overestimating the risk of myocardial infarction in men.
\end{itemize}

\subsubsection{Results}

As in the few-shot experiment, results are presented in Tables \ref{tab:skew_baseline} for non-LLM methods and \ref{tab:skew_llm} for LLM-based methods. LAAT effectively mitigates dataset biases and achieves the highest performance across all datasets, surpassing non-LLM baselines by more than 20 ROC AUC points in some cases. Although FeatLLM outperforms traditional baselines, it achieves shared top performance with LAAT on only two datasets and lags significantly behind LAAT models on the remaining datasets. The superior performance of LAAT on this challenging benchmark further highlights the benefits of attribution guidance — not only in enhancing performance in data-scarce scenarios but also in addressing subtle, hard-to-detect dataset biases that could otherwise compromise the integrity of the machine learning pipeline.

\begin{table}
\centering
\caption{ROC AUC scores of baseline methods on the biased dataset experiments. Best scores for each dataset, across both LLM and non-LLM approaches, are emphasized in \textbf{bold}. Multiple bolded values indicate that their differences were not statistically significant according to the Wilcoxon signed-rank test at $p=0.05$.}\label{tab:skew_baseline}
\begin{tabular}{l||l|l|l|l|l|l}
\hline
\textbf{Model} & LR & MLP & RF & XGB & CatBoost & TabPFN \\
\textbf{Dataset} &  &  &  &  &  &  \\
\hline
\hline
bodyfat & $84.5_{4.4}$ & $81.1_{4.7}$ & $78.1_{6.2}$ & $75.0_{7.3}$ & $82.3_{4.1}$ & $85.0_{3.9}$ \\
\cline{1-7}
breast-ljub & $54.3_{6.6}$ & $54.4_{5.4}$ & $55.8_{7.4}$ & $54.6_{9.5}$ & $54.0_{6.6}$ & $52.9_{7.6}$ \\
\cline{1-7}
cdc-diabetes & $62.2_{0.4}$ & $62.4_{0.3}$ & $60.3_{2.3}$ & $49.3_{1.3}$ & $60.2_{2.1}$ & $50.1_{1.3}$ \\
\cline{1-7}
contraceptive & $57.3_{3.0}$ & $50.9_{5.8}$ & $55.7_{5.3}$ & $47.6_{2.2}$ & $57.0_{2.8}$ & $50.1_{2.9}$ \\
\cline{1-7}
diabetes & $75.8_{3.3}$ & $72.9_{4.1}$ & $72.9_{4.6}$ & $60.7_{8.8}$ & $75.0_{3.2}$ & $76.4_{3.1}$ \\
\cline{1-7}
indian-liver & $60.4_{6.7}$ & $56.8_{5.3}$ & $54.9_{8.4}$ & $49.3_{6.1}$ & $58.9_{5.6}$ & $59.8_{5.5}$ \\
\cline{1-7}
myocardial & $55.1_{8.2}$ & $55.8_{5.9}$ & $58.5_{5.1}$ & $54.6_{8.7}$ & $55.6_{9.9}$ & $50.0_{7.8}$ \\
\cline{1-7}
\hline
\end{tabular}
\end{table}

\begin{table}
\centering
\caption{ROC AUC scores of LLM-based models on the biased dataset experiments. Best scores for each dataset, across both LLM and non-LLM approaches, are emphasized in \textbf{bold}. Multiple bolded values indicate that their differences were not statistically significant according to the Wilcoxon signed-rank test at $p=0.05$.}\label{tab:skew_llm}
\begin{tabular}{l||ll|lll|lll}
\hline
\textbf{Method} & \multicolumn{2}{c|}{FeatLLM} & \multicolumn{6}{c}{LAAT} \\
\hline
\textbf{Model} & \multicolumn{2}{c|}{Ensemble} & \multicolumn{3}{c|}{LR} & \multicolumn{3}{c}{MLP} \\
\hline
\textbf{Dataset} & $Gem_{2.0}$ & $G_{4om}$ & $LLa_{3.3}$ & $Gem_{2.0}$ & $G_{4om}$ & $LLa_{3.3}$ & $Gem_{2.0}$ & $G_{4om}$  \\
\hline
\hline
bodyfat & $80.9_{6.4}$ & $79.3_{6.2}$ & $84.5_{16.2}$ & $83.4_{18.0}$ & $84.3_{17.0}$ & $89.5_{4.2}$ & $\mathbf{91.0_{3.6}}$ & $90.4_{4.0}$ \\
\cline{1-9}
breast-ljub & $67.8_{8.7}$ & $62.9_{7.2}$ & $\mathbf{71.1_{9.6}}$ & $71.1_{8.6}$ & $70.5_{8.7}$ & $\mathbf{73.6_{6.9}}$ & $\mathbf{73.5_{6.8}}$ & $\mathbf{73.2_{6.9}}$ \\
\cline{1-9}
cdc-diabetes & $75.4_{3.3}$ & $76.1_{0.8}$ & $72.5_{0.3}$ & $\mathbf{79.0_{0.2}}$ & $75.6_{0.3}$ & $72.4_{0.3}$ & $78.9_{0.3}$ & $74.1_{0.9}$ \\
\cline{1-9}
contraceptive & $54.8_{8.6}$ & $51.8_{3.9}$ & $63.9_{3.4}$ & $\mathbf{66.6_{2.2}}$ & $62.9_{3.4}$ & $63.9_{3.4}$ & $66.1_{2.2}$ & $62.9_{3.5}$ \\
\cline{1-9}
diabetes & $76.8_{3.8}$ & $76.3_{2.8}$ & $78.8_{6.9}$ & $78.4_{7.2}$ & $78.6_{7.3}$ & $79.9_{4.1}$ & $\mathbf{80.0_{3.9}}$ & $\mathbf{80.2_{4.0}}$ \\
\cline{1-9}
indian-liver & $67.4_{6.4}$ & $\mathbf{69.3_{6.1}}$ & $\mathbf{71.1_{5.7}}$ & $\mathbf{71.4_{5.5}}$ & $\mathbf{69.3_{5.7}}$ & $\mathbf{72.1_{4.3}}$ & $\mathbf{72.1_{3.9}}$ & $71.2_{4.5}$ \\
\cline{1-9}
myocardial & $54.0_{9.2}$ & $\mathbf{62.5_{5.1}}$ & $63.8_{6.2}$ & $\mathbf{66.2_{6.4}}$ & $64.4_{5.6}$ & $59.4_{7.2}$ & $\mathbf{64.8_{7.3}}$ & $61.6_{7.3}$ \\
\cline{1-9}
\hline
\end{tabular}
\end{table}


\subsection{Hyperparameter Analysis}

\begin{figure}[!htb]
   \begin{minipage}{0.48\textwidth}
     \centering
     \includegraphics[width=\linewidth]{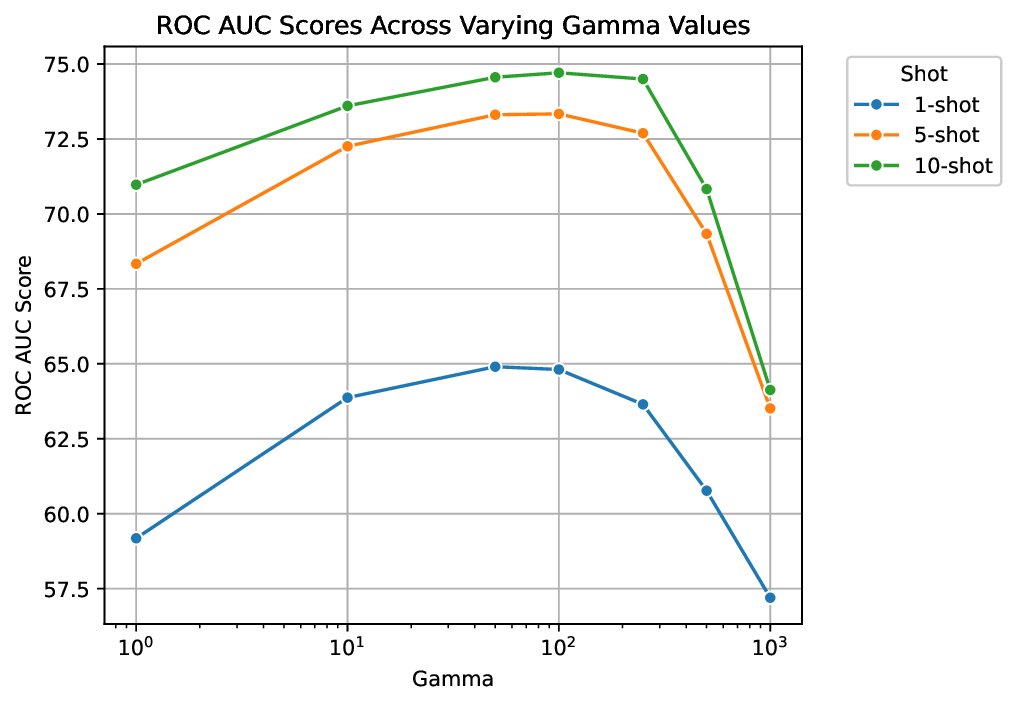}
     \caption{ROC AUC scores of LAAT aligned logistic regression over varying gamma values.}\label{fig:gamma_sweep}
   \end{minipage}\hfill
   \begin{minipage}{0.48\textwidth}
     \centering
     \includegraphics[width=\linewidth]{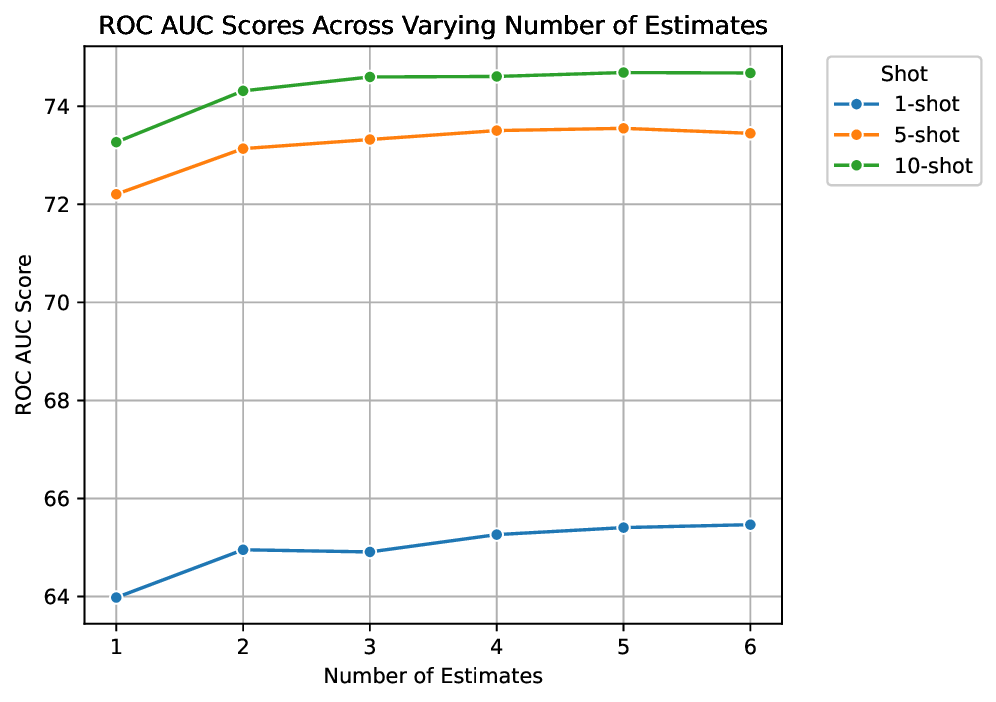}
     \caption{ROC AUC scores of LAAT aligned logistic regression over varying number of estimates.}\label{fig:nestimates_sweep}
   \end{minipage}
\end{figure}

We investigate the impact of varying the regularization factor \(\gamma\) and the number of importance score estimations on the final performance of LAAT models. Specifically, we assess the performance of a logistic regression model trained using LAAT alignment with importance scores provided by GPT-4o mini. The experimental results are presented in Figures \ref{fig:gamma_sweep} and \ref{fig:nestimates_sweep}, corresponding to the $\gamma$ variation and the number of importance score estimations, respectively.  

Our findings indicate that increasing $\gamma$ initially enhances model performance, reaching an optimal value around $\gamma = 100$. However, further increases beyond $\gamma = 250$ lead to a substantial decline in performance. In contrast, increasing the number of importance score estimations results in a steady performance improvement, which plateaus at approximately four to five estimations, depending on the k-shot setting. These results suggest that while additional score generation and ensembling can enhance performance, their benefits diminish beyond a certain threshold.

\section{Limitations}
While LAAT offers significant benefits, it has several limitations. First, the method is restricted to tabular data and supports only binary classification tasks. Additionally, LAAT requires features to be describable in natural language, limiting its applicability to datasets with anonymized or uninterpretable features. Finally, its effectiveness depends on the general knowledge embedded in LLMs, which may vary across different data domains.

\section{Conclusion}
We introduce \textbf{L}arge Language Model \textbf{A}ttribution \textbf{A}ligned \textbf{T}raining (\textbf{LAAT}), a novel approach that leverages importance scores inferred by large language models (LLMs) as local attribution guides. This method effectively harnesses the generalization capabilities of LLMs while employing simple, traditional machine learning models. Although gradient boosting trees have traditionally dominated tabular data learning, LAAT significantly outperforms existing methods in few-shot learning and biased dataset scenarios. Notably, it surpasses FeatLLM, the current state-of-the-art LLM-based few-shot learning approach.

Beyond its standalone effectiveness, LAAT is highly versatile and can be used in conjunction with more complex models, provided they are differentiable — for example, TabPFN. Additionally, it can be combined with feature preprocessing techniques such as FeatLLM. Given the interpretability of LLM-derived importance scores, LAAT could be incorporated into interactive, chat-based interfaces, enabling human experts to refine these scores before finalizing them. This human-in-the-loop refinement could further enhance model performance by leveraging domain expertise.

While we focus on input gradient-based attribution in this work, future research could explore alternative attribution methods to further enhance LAAT’s effectiveness. Additionally, future research could explore extending LAAT to multiclass classification problems by generating LLM-based feature attribution vectors for each class. Subsequent training could regularize local attributions based on the class of the current example. Similarly, in regression tasks, the score generation prompt of LAAT could be adapted to accommodate continuous target variables.

Furthermore, LAAT’s framework could be expanded beyond tabular data to other modalities, such as images, by integrating it into concept bottleneck models \cite{DBLP:conf/icml/KohNTMPKL20}. These architectures produce high-level, human-interpretable features that could serve as input to a LAAT-augmented classifier, facilitating improved feature selection and attribution in vision tasks. Such an extension could enhance both interpretability and generalization in image-based applications. We leave the exploration of these research directions to future work.

\begin{credits}
\subsubsection{\discintname}
The authors have no competing interests to declare that are relevant to the content of this article.
\end{credits}
%
%
%
\bibliographystyle{splncs04}
\bibliography{mybibliography}






\appendix
\clearpage

\section*{Appendix}
We provide dataset details, task descriptions, score extraction prompt, hyperparameter search spaces, and loss landscapes used in our experiments.


\begin{table}
\centering
\caption{Basic information about dataset used in our experiments.}\label{tab:basic_dataset}
\begin{tabular}{l||r|r|r|r}
\hline
Dataset & Samples & Num. features & Cat. features & Pos. label ratio \\
\hline
\hline
\textbf{adult} & 48842 & 14 & 8 & 23.93\% \\
\textbf{bank} & 45211 & 16 & 9 & 11.70\% \\
\textbf{breast-ljub} & 277 & 16 & 0 & 29.24\% \\
\textbf{cdc-diabetes} & 253680 & 21 & 0 & 13.93\% \\
\textbf{diabetes} & 768 & 8 & 0 & 34.90\% \\
\textbf{electricity} & 45312 & 8 & 0 & 42.45\% \\
\textbf{myocardial} & 686 & 91 & 13 & 22.16\% \\
\hline
\end{tabular}
\end{table}


\begin{table}
\centering
\caption{Task descriptions utilized in the experiments.}\label{tab:task_descriptions}
\begin{tabular}{l||p{0.8\textwidth}}
\hline
Dataset &  Task Description \\
\hline
\hline
\textbf{adult} &  Predict whether this person earns more than 50000 dollars per year. Yes or no?\\
\hline
\textbf{bank} & Predict whether this client will subscribe to a term deposit. Yes or no? \\
\hline
\textbf{breast-ljub} &  Predict whether this patient's breast cancer will reoccur. Yes or no?\\
\hline
\textbf{cdc-diabetes} & Predict whether the patient has diabetes. Yes or no?\\
\hline
\textbf{diabetes} & Predict whether the patient has diabetes. Yes or no?\\
\hline
\textbf{electricity} & In the electricity market, prices are not fixed and are affected by demand and supply of the market. They are set every five minutes. Electricity transfers in the state A to/from the neighboring state B are done to alleviate fluctuations. Based on the current measurement, predict whether the price of electricity in state A will go up. Yes or no?\\
\hline
\textbf{myocardial} & Predict whether the myocardial infarction complications data of this patient show chronic heart failure. Yes or no?\\
\hline
\end{tabular}
\end{table}


\begin{lstlisting}[caption={Prompt template for LLM importance score generation.},label=prompt:score_generation]

You are an expert at assigning importance scores to features
used for a classification task. For each feature, output an
integer importance score between -10 and 10. Positive scores
suggest that an increase in the feature's value boosts the
class probability, whereas negative scores indicate that an
increase in the feature's value reduces the class probability.
You have to include a score for every feature.
Task: {task_prompt}
Features:
{features_prompt}
Output the importance scores for the class "{label}".

Think step by step and output an integer importance score
between -10 and 10 for each feature. You must specify each
feature individually, in order of its appearance.
\end{lstlisting}


\begin{table}
\caption{Hyperparameter search spaces for baseline models.}\label{tab:hyperparameters}
\begin{tabular}{l||p{0.8\textwidth}}
\hline
Model & Hyperparameter search space \\
\hline
\hline
\textbf{Logistic Regression} &  C: [100, 10, 1, 1e-1, 1e-2, 1e-3, 1e-4, 1e-5]\\
\hline
\textbf{MLP} & alpha: [0.001, 0.01, 0.1, 1, 10] \newline learning\_rate\_init: [0.1, 0.01, 0.001, 0.0001] \\
\hline
\textbf{XGBoost} &  max\_depth: [2, 4, 6, 8, 10] \newline
       alpha: [1e-4, 1e-3, 1e-2, 1e-1, 1, 10] \newline
        lambda: [1e-4, 1e-3, 1e-2, 1e-1, 1, 10] \newline
        eta: [0.01, 0.03, 0.1, 0.3]\\
\hline
\textbf{Random Forest} & bootstrap: [True, False] \newline
        max\_depth: [2, 4, 6, 8, 10] \newline
        n\_estimators: [2, 4, 8, 16, 32, 64]\\
\hline
\textbf{CatBoost} & colsample\_bylevel: [0.01, 0.03, 0.06, 0.1] \newline
        boosting\_type: ["Ordered", "Plain"] \newline
        depth: [2, 4, 6, 8, 10]\\
\hline
\end{tabular}
\end{table}


\begin{figure}
    \centering
    \begin{subfigure}{0.49\textwidth}
        \centering
        \includegraphics[width=\linewidth]{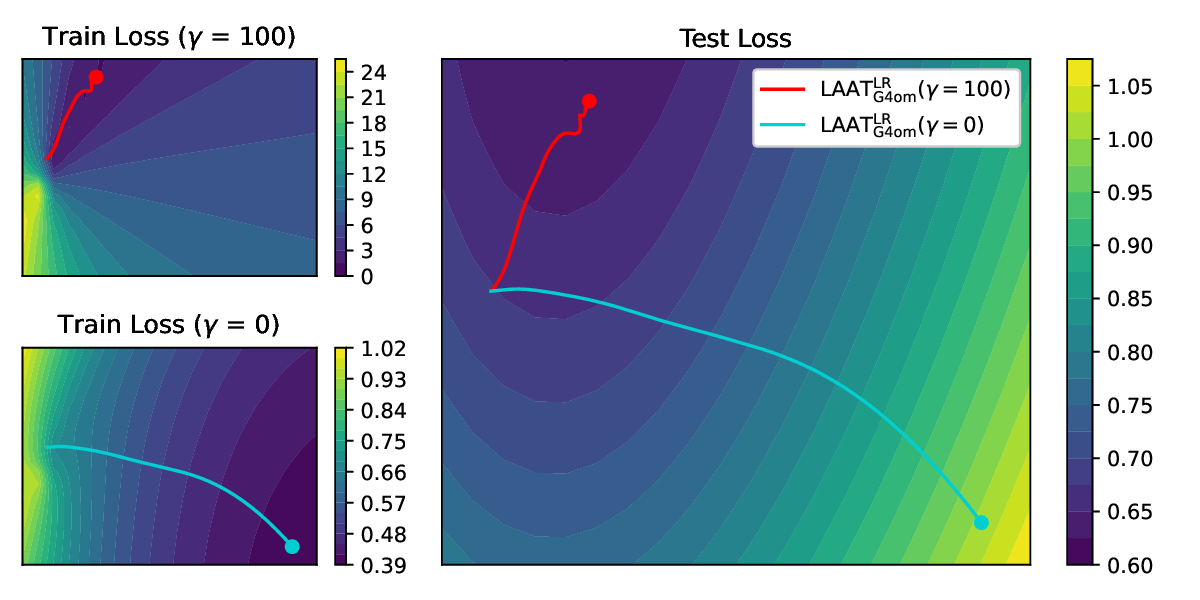}
        \caption{Loss landscapes - \textbf{breast-ljub} dataset.}
    \end{subfigure}
    \hfill
    \begin{subfigure}{0.49\textwidth}
        \centering
        \includegraphics[width=\linewidth]{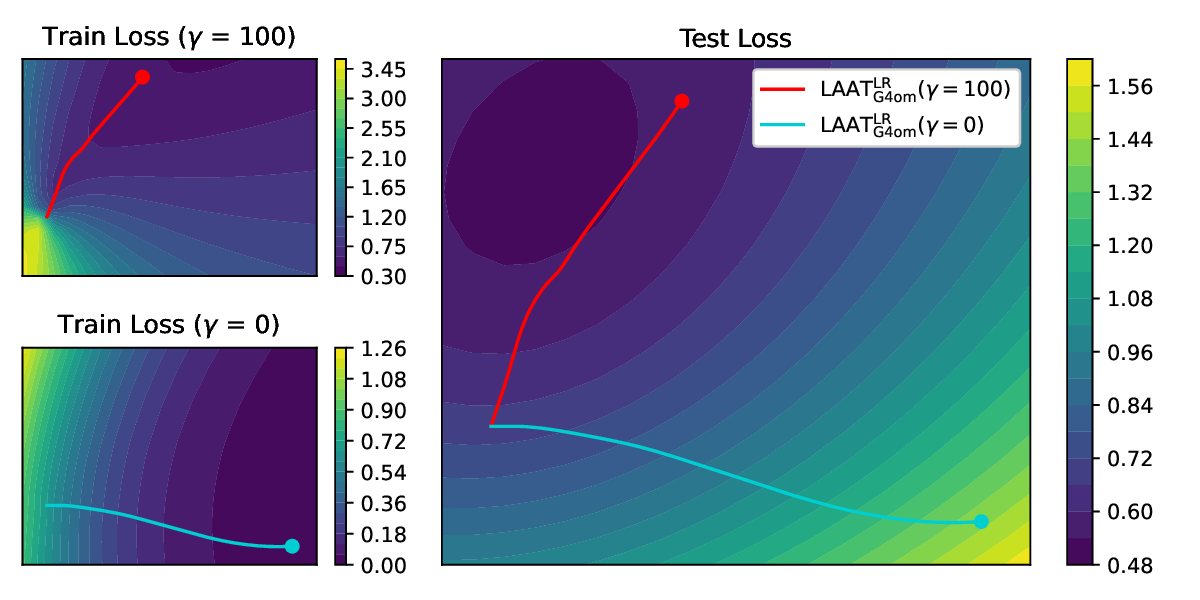}
        \caption{Loss landscapes - \textbf{myocardial} dataset.}
    \end{subfigure}
    \caption{Loss landscapes for logistic regression on the \textbf{breast-ljub} and \textbf{myocardial} datasets, comparing models with ($\gamma = 100$, top left) and without ($\gamma = 0$, bottom left) attribution alignment loss. The addition of attribution alignment loss better aligns training (left) and test (right) loss landscapes, guiding the model towards minima that improve generalization.}
    \label{fig:loss_landscapes_breast_myocardial}
\end{figure}

\end{document}